\documentclass{amsart}

\usepackage{url}
\usepackage{amsmath,amsfonts}
\usepackage{latexsym}
\usepackage[english]{babel}
\usepackage{graphicx}

\graphicspath{{img/}}

\newcommand{\E}{\operatorname{E}}
\newcommand{\IR}{\mathbb{R}}
\newcommand{\purity}{\operatorname{purity}}

\title{Entropy methods for the confidence assessment of probabilistic
classification models}

\author{Gabriele Nunzio Tornetta}
\email{gabriele.n.tornetta@gmail.com}
\address{Edinburgh, UK}

\begin{document}

\maketitle

\begin{abstract}%
Many classification models produce a probability distribution as the outcome of
a prediction. This information is generally compressed down to the single class
with the highest associated probability. In this paper we argue that part of the
information that is discarded in this process can be in fact used to further
evaluate the goodness of models, and in particular the confidence with which
each prediction is made. As an application of the ideas presented in this paper,
we provide a theoretical explanation of a confidence degradation phenomenon
observed in the complement approach to the (Bernoulli) Na\"ive Bayes generative
model.
\end{abstract}

\section{Introduction}

A classification model, or \emph{classifier} in brief, is a statistical model
that produces a qualitative (that is, discrete) output. Within the scopes of
supervised learning, its target values are generally known \emph{a-priori} and
are sometimes referred to as \emph{classes}. As opposed to regression, it is
often the case that classification problems require \emph{ad-hoc}, and generally
not uniquely agreed-upon, definitions of certain concepts. Typical examples that
come to mind are the Bias-Variance decomposition and the Prediction Error
\cite{tibshirani96}.

A similar problem arises when one looks at evaluation metrics for classifiers.
Many different evaluation techniques have been developed that apply only to
classification problems. One of the most widely used is certainly the
\emph{confusion matrix}, together with the related scores that can be computed
from it, like \emph{recall}, \emph{precision}, \emph{accuracy} \cite{stehman97}
to name a few. Receiver Operating Characteristics (ROC) Analysis, which began
with electrical and radar engineers for martial purposes, has been introduced
in Machine Learning problems by \cite{spackman89} in 1989 and constitutes now a
standard evaluation tool for classification models. A good survey on the topic
can be found in \cite{fawcett06}.

Many classification models, though, provide an interim probability distribution
$p$, or a score function that can be regarded as a probability distribution,
over the set of allowed classes. This is done either directly or via ensemble
techniques, like bagging \cite{domingos99}, and the final, discrete output is
produced by picking the argument with the highest probability, or score. We
shall refer to models of this kind as \emph{probabilistic}, as opposed to the
purely \emph{discrete} cases that only provide a class but no interim
probability distribution or score function.

Now, it should be evident that the process of computing $\arg\max p$, whilst
yielding the result of interest, which in many cases is just the predicted
class, is also discarding the further (potentially useful) information encoded
in the full description of the probability distribution $p$.

An interesting question is whether the extra information that resides in the
distribution $p$ is, in some ways, useful or not. Our aim here is to argue that,
whilst a classification model, as previously stated, is generally asked to
provide a choice from a discrete set of classes, any such $p$ can provide
information that can be used to assess how \emph{confident} the model is.

The aim of this paper is to formally define and analyse new evaluation metrics
for probabilistic classification models that can complement the information
provided by the standard evaluation metrics and techniques mentioned earlier. To
illustrate the theoretical importance of such metrics, we illustrate an
application where we provide a theoretical explanation of why the probability
distributions produced by the Complement Na\"ive Bayes model of \cite{rennie}
are observed to be, in many cases, closer to the uniform distribution than those
generated by, e.g., the traditional Bernoulli Na\"ive Bayes classifier. From a
practical point of view, we provide text classification experiments with which
we reproduce the phenomenon and show how the new metrics capture it.

\section{Confidence scores}

In order to avoid confusion with well established terminology, we shall start by
clarifying that, throughout this paper, the use of the word \emph{confidence} is
by all means not related to the concept of confidence intervals. By the
confidence of a probabilistic classifier we mean the value of the highest
predicted probability. For example, if two binary classifiers provide the
predictions $(0.45, 0.55)$ and $(0.1, 0.9)$ respectively, we say that the second
model is more confident in its prediction, as $0.9 > 0.55$. Whether the
prediction is correct or not is a different question. In this sense, our meaning
of confidence is somehow related to a confidence interval, but that is perhaps
as far as the analogy goes.

Thus, when comparing two classifiers, say $h$ and $g$, we may say that the
predictions of $h$ are \emph{sharper} than those of $g$ if the confidence of $h$
is higher, on average, than that of $g$. The rest of this section is devoted to
making the notion of \emph{sharpness} more rigorous and to providing evaluation
metrics by which one can get an idea of the confidence of a probabilistic
classification model.

Some other authors prefer to use the antipodal concept of \emph{uncertainty}
when it comes to assessing probabilistic classifiers. For example, notions
similar to the ones presented in this paper, based on entropy (or information)
can be found in \cite{zhang2019}. The use of \emph{deviance} as a measure of the
uncertainty of probabilistic classifiers appears to be common practice (see,
e.g., \cite{ritschard06} for a discussion on decision trees). Uncertainty can
also be quantified in terms of probability, as argued in \cite{schetinin04},
where the notion of \emph{Uncertainty Envelope Technique} is introduced.

In this paper, however, we present and focus on a measure that is bounded in
nature within the interval $[0, 1]$ and that correlates with the classifier
confidence, that is, the higher the score the more confident the model is. We
will then exploit some of the basic properties of entropy to give a mathematical
justification of a phenomenon of decreased confidence in certain Na\"ive Bayes
models with good training scores.

\subsection{Entropy score}

Given a probability distribution $p$ on a discrete set $C$ of finite cardinality
$|C|=n$, we may compute its entropy, that is, the number
  $$H[p] = -\sum_{c\in C}p(c)\log p(c).$$
It is well known and immediate to verify that $H$ attains its minimum value of
$0$ when $p(c)=1$ for some $c\in C$, and its maximum value of $\log n$ when
$p(c)=\frac1n$ for any $c\in C$.

Suppose now that we have two distributions, $p$ and $q$, over the same space
$C$, and such that $\arg\max p = \arg\max q$, but with $H[p] < H[q]$. If $p$ and
$q$ come from two classification models for the same problem, we now appreciate
that, whilst the predicted classes are the same, the model that yields the
distribution $p$ is doing so with a higher \emph{confidence} in its ``choices''
of probable classes. If we find that the two models have comparable training
scores and we are to choose one, how can we use an ``entropy score'' as a
tiebreaker? Of course, if the training scores are quite low, we would hope for
such entropy score to be poor too, for otherwise we would have models that are
very confident in making the wrong predictions. This is probably a further
indication that perhaps we should look for models from a different family. If we
then disregard this possibility and only consider models with good training
scores, we can argue that the preference should be given to the model that
yields sharper distributions, i.e. that with the highest entropy score.

Based on the argument above, one could then think of defining the mentioned
\emph{entropy score} for model evaluation as the $[0,1]$-valued metric
  \begin{equation}\label{eq:entropy_score}
    h = 1 - \frac{\E\left[H[p]\right]}{\log n}.
  \end{equation}
The closer $h$ is to 1, the \emph{sharper} the distributions $p$ are on average,
and hence the more confident the classification model. Of course, as we have
just seen from the above discussion, this metric alone is not enough for model
selection, as a model could be very confident in making wrong predictions, but
can certainly be used when trying to decide between models that otherwise seem
to perform equally well when evaluated against other metrics.

\subsection{Purity}

Another possible way of assessing the confidence of a classification model that
also takes into account the accuracy of its predictions is by looking at the
\emph{probabilistic} confusion matrix \cite{cen}
  $$P_{ij} = \frac1{|T_i|}\displaystyle\sum_{s\in T_i} p_s(j)$$
where $T_i$ is the subset of the sample set $T$ with true classes $i\in C$ and
$p_s$ is the probability distribution that the model produced for sample $s$.

We shall say that a probabilistic confusion matrix is \emph{pure} if it
coincides with the identity matrix $\delta_{ij}$. It goes without saying that,
in reality, almost no probability distribution matrix will ever be pure, as
$\{I\}$ is just a subset of null (Lebesgue) measure inside the space of all the
square matrices $\operatorname{Mat}_{n\times n}$.
Nonetheless we can introduce a notion of \emph{purity}, which gives a measure of
how close the matrix $P$ is to the ideal case, i.e. the identity matrix $I$. A
possible definition is to treat both $P$ and $I$ as elements of $\IR^{n^2}$ and
take the (normalised) Euclidean norm, viz.
  $$\purity(P) := 1 - \frac{\Vert P-I\Vert_2}{\sqrt{2n}}.$$
Hence, with this definition, $\purity(I) = 1$, whereas for any
other matrix $P$ we would have a purity in $[0,1)$.

To get a feel for this metric and how it can assess the confidence of a
probabilistic classifier, let us look at the case $n = 2$. If the predictions
are accurate and confident, we expect to find a probabilistic confusion matrix
that is very close to the identity matrix. Hence, we would expect a purity quite
close to 1. On the other hand, for a model that gives the wrong answers with
great confidence, we expect to find a matrix $P$ very close to
  $$\sigma_1 = \begin{bmatrix}0&1\\1&0\end{bmatrix}.$$
Hence, the purity of $P$ would be close to
  $$\purity(\sigma_1) = 0.$$
The case of a classifier that produces predictions with a distribution quite
close to the uniform one would give a matrix $P$ close to $\frac12(I+\sigma_1)$,
for which we have
  $$\purity\left(\frac12(I+\sigma_1)\right) = \frac12.$$
Therefore, a classifier that exhibits poor confidence in its prediction would
have an intermediate value of purity for the probabilistic confusion matrix.
Observe that, in the general case of dimension $n$, we would have a purity value
of $\sqrt{\frac{n-1}{2n}}$, which tends to $\frac1{\sqrt2}$ from below as $n$
grows arbitrarily large.

To summarise, from the above observations we have learned that a purity close to
the extreme values of the interval $[0, 1]$ indicate a model that is making
pretty confident predictions, which are good when the value is close to 1 and
bad when it is close to 0. Intermediate values are, in general, the indication
of a model that is somewhat \emph{uncertain} about its predictions, regardless
of whether they are good or bad.

\section{\label{sec:applications}Applications}

We shall now apply the entropy considerations of the previous section to the
Complement Na\"ive Bayes model described in \cite{rennie}. In fact, here we
shall consider a variant where even the \emph{a-priori} class probabilities are
complemented. The result that we obtain here can also be observed in experiments
carried out with the model defined in \cite{rennie}.

\subsection{The complement assumption} First of all, we shall briefly explain
why one might want to destroy the generative properties of the standard Na\"ive
Bayes model by manually tweaking its parameters. As argued in \cite{rennie},
when one is dealing with a heavily unbalanced sample set, the traditional
Na\"ive Bayes model tends to favour classes with a larger support during
training. Indeed it is hard for a model to understand well the ``meaning'' of a
class when it can only see just a few example, and it is therefore quite likely
to lean towards better understood ones.

This ``bias'' can be alleviated by complementing on classes, that is, by
considering the contributions from the features that appear in samples from
classes other than a certain class $c$. In a sense, we are now asking the
classification model to recognise a class by learning what the class is
\emph{not}, in relation to a given set of classes. This way, the model can have
access to a larger number of examples for each class, making the training set
less unbalanced.

Taking a step back, we start by looking at the generative Bayesian approach,
whereby the probability of having class $c$ \emph{given} the feature vector
$x\in\{0,1\}^m$, can be expressed, up to a normalisation factor, as
  $$p(c|x) \propto p(x|c) p(c).$$
The na\"ive assumption takes the form of conditional independence for the
marginals of each component of $x$, namely
  $$p(x|c) = \prod_{\mu=1}^m p(x_\mu|c).$$
The (Bernoulli) Na\"ive Bayes model is then characterised by the parameters
  $$\phi_{\mu c} = p(x_\mu=1|c)\quad\text{and}\quad\psi_c = p(c),$$
with the obvious constraint
  $$\sum_{c\in C}\psi_c = 1.$$
When the (log-)likelihood is maximised over a given training set $T$, we obtain
the estimates
  \begin{equation}\label{eq:params}
    \phi_{\mu c} = \frac{N_{\mu c}}{N_c}
      \quad\text{and}\quad
    \psi_c = \frac{N_c}N,
  \end{equation}
where $N$ is the cardinality of $T$, $N_c$ the number of samples of true class
$c$ and $N_{\mu c}$ the number of samples in $T$ where the $\mu$th feature is 1
and is of true class $c$.

To make the argument slightly more concrete, suppose that the problem at hand is
text classification. Each $x_\mu$ indicates the presence or absence of the word
$\mu$ in  the document $x$. We then find that the probability of being of class
$c$ for a document containing the word $\mu$ can be estimated with
  $$p(c|x_\mu=1) = \frac{N_{\mu c}}{N_\mu},$$
where $N_\mu$ is the number of documents in the training set containing at least
one occurrence of the word $\mu$, viz.
  $$N_\mu = \sum_{c\in C}N_{\mu c}.$$

We now apply the ideas set out in \cite{rennie} to the estimates
\eqref{eq:params} to produce a ``complement'' model. We set
\begin{equation}\label{eq:params}
  \tilde\phi_{\mu c} = \frac{N - N_c}{N_\mu - N_{\mu c}}
    \quad\text{and}\quad
  \tilde\psi_c = \frac N{N - N_c},
\end{equation}
with which we get
  $$q(c|x_\mu = 1) \propto \frac1{1-p(c|x_\mu=1)}.$$
It is important now to make the observation that, when $n=2$, $\tilde\phi_{\mu
c}=\phi_{\mu c}$ and $\tilde\psi_c = \psi_c$, and therefore, from now on, we
make the assumption that $n > 2$. That is, we are only interested in multi-class
classification problems, where the results that follow are non-trivial.

\subsection{Degraded confidence} As shown in the already cited work
\cite{rennie}, the complement construction can alleviate the problems caused by
unbalanced classes. Unfortunately, as we shall now see, this comes at a cost.

\begin{figure}[tp!]
    \begin{center}
    \scalebox{.65}{\input{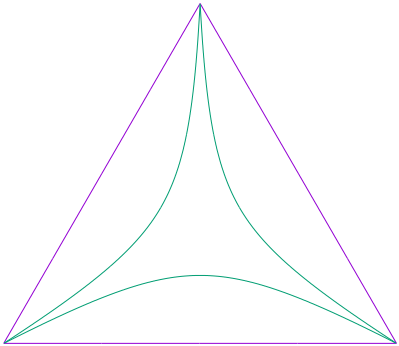}}
    \end{center}
    \caption{\label{fig:transform}Representation of the map \eqref{eq:transform}
    for the case $n=3$ on the $q_1 + q_2 + q_3 = 1$ plane in $\mathbb R^3$,
    isometrically projected down to $\mathbb R^2$. The 2-simplex $\Delta^2$
    (purple) is shrunk to a star-shaped domain (green) that seems to ``implode''
    towards the uniform distribution, which is the centre of mass of both
    $\Delta^2$ and the image of the transformation.}
  \end{figure}

Abstracting from the result obtained at the end of the previous section, we
shall now focus on the transformation $q:\Delta^{n-1}\to\Delta^{n-1}$, where
$\Delta^{n-1}$ is the standard $(n-1)$-simplex, given by
  \begin{equation}\label{eq:transform}
    q_k(p) = \frac{\frac1{1-p_k}}{\sum_{k=1}^n\frac1{1-p_k}},
      \quad k =1,\ldots,n.
  \end{equation}
It is easy to see (cf. Figure \ref{fig:transform}) that this map moves every
probability distribution
$p\in\Delta^{n-1}$ closer to the uniform distribution
$(\frac1n,\ldots,\frac1n)$, which is a fixed point, with the exception of all
the vertices, which are also fixed points of the transformation. As a direct
consequence of this property, we conclude immediately that
  $$H[q(p)] \geq H[p],\quad\forall p\in\Delta^{n-1},$$
with the equality attained only on the fixed points. The larger $n$, the closer
some probability distributions are moved towards the uniform case. Consider, for
instance, the image of the distribution
  $$\left(\frac12, \frac12, 0, \ldots, 0\right),$$
which is
  $$\left(\frac2{n+2}, \frac2{n+2}, \frac1{n+2}, \ldots, \frac1{n+2}\right),$$
quite evidently asymptotic to the uniform distribution as $n\to\infty$.
Based on the previous discussion on entropy, we further conclude that the
confidence of the prediction that comes with the probability $q$ is lower than
the confidence associated with $p$. We then expect the probability distributions
generated by the Complement Na\"ive Bayes model to be certainly more accurate,
but less sharp than the standard Na\"ive Bayes case. In a sense, the complement
construction is trading in some of the model's confidence for extra accuracy.

\subsection{Experimental results}

We shall now give some experimental evidence to the claim that was made at the
beginning, namely, that the confidence degradation that has been proved for the
complement model described in this paper can also be observed in the Complement
Na\"ive Bayes model of \cite{rennie}.

To this end, we have prepared two different experiments, where we compare three
different Na\"ive Bayes models, namely Binomial, Multinomial and Complement
Na\"ive Bayes, as implemented in scikit-learn \cite{scikit-learn}. For the
Binomial model we have used a binary TfidfTransformer layer with no inverse
document frequency, whereas for the other two models we have used the default
parameters. All the metrics are evaluated out-of-sample on a separate test set.

The first experiment is based on the 20
Newsgroup\footnote{http://qwone.com/\textasciitilde{}jason/20Newsgroups/} data
set available via scikit-learn. We have selected three news categories, namely
\texttt{soc.religion.christian}, \texttt{comp.graphics}, \texttt{sci.med}, and
adjusted the class supports to produce both a balanced and unbalanced training
data set.

\begin{figure}[t!]
  \begin{center}\small
  \scalebox{1}{\input{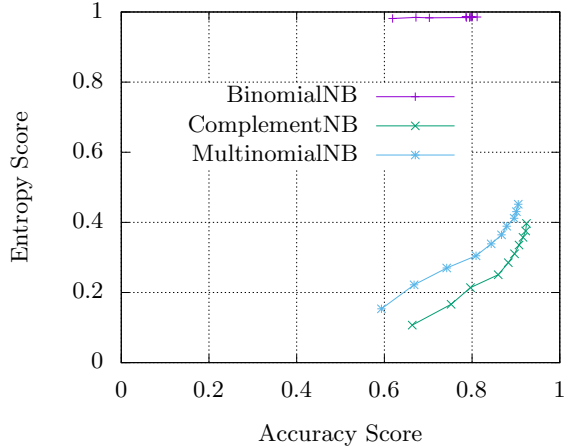}}
  \end{center}
  \caption{\label{ng20_balanced}The accuracy \emph{versus} entropy score for the
  20 Newsgroup experiment with balanced supports. The left-most points are
  obtained from models trained with just the 10\% of all the samples in each
  class. Supports are increased in a linear fashion up to 100\% of all the
  samples in each class. As more data is fed into the models, their accuracy
  increases. However, the confidence of the complement model stays below the
  other curves.}
\end{figure}

Figure \ref{ng20_balanced} is the plot of accuracy \emph{versus} entropy score
for the balanced case. The different points have been obtained by increasing the
support of all the classes simultaneously by the same factor of the total
support. As expected, the accuracy increases with the supports, and so the
points move from left to right as more data is fed to the models.

We see that the complement model starts with the highest accuracy with the
lowest support, but confidence is poor. As we let the supports grow, accuracy
grows too and outperform the other models with the same support, but we can also
see that the confidence of the complement model stays consistently below the
other curves, as it was expected.

\begin{figure}[t!]
  \begin{center}\small
  \scalebox{1}{\input{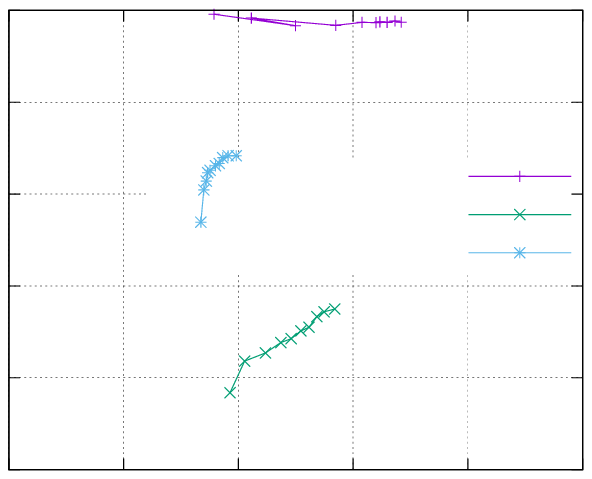}}
  \end{center}
  \caption{\label{ng20_unbalanced}The accuracy \emph{versus} entropy score for
  the 20 Newsgroup experiment with unbalanced supports. As in the balanced case,
  we increase the support size of each class linearly, but we keep relative
  proportions of roughly $2:5:10$ among the three classes mentioned in the
  paper. The picture that we get is quite different from the balanced case, but
  we still spot the poor confidence of the complement model.}
\end{figure}

Figure \ref{ng20_unbalanced} shows similar trends for the unbalanced case.
Indeed, with low support, the complement model gives the highest accuracy, but
confidence remains poor when compared against the other models. In this
particular instance, however, the Binomial model seems to provide better
accuracy and confidence scores as the supports increase.

The second experiment that we present is different from the first one, in the
sense that we have not fixed the classes, but we have set a threshold for the
class supports. As a consequence, it is more difficult to compare points on the
same curve with each other, but we can still spot the expected behaviour of the
complement model.

\begin{figure}[t!]
  \begin{center}\small
  \scalebox{1}{\input{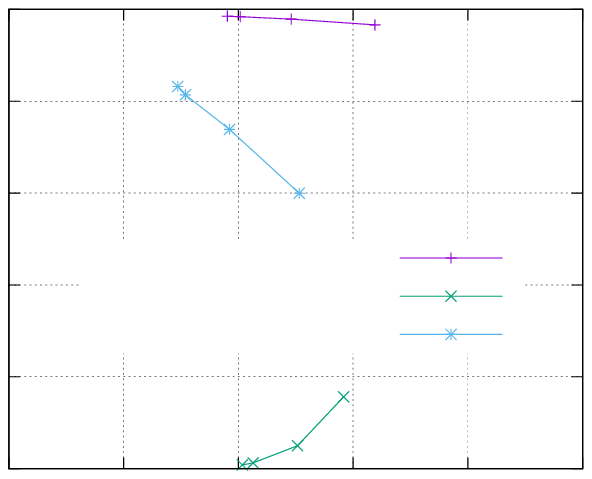}}
  \end{center}
  \caption{\label{wiki}The accuracy \emph{versus} entropy score for the
  Wikipedia Movie Plots experiment. The threshold values that have been used are
  100, 200, 500, 1000. The corresponding numbers of classes that were selected
  from the training set are 29, 21, 9, 4. Low threshold values then imply more
  classes and hence lower accuracy (left-ward in the plot). A higher threshold
  filters out many more classes, thus allowing the models to focus on a
  restricted pool of choices, yielding better accuracy (rightward in the plot).}
\end{figure}

The training data is based on the Wikipedia Movie Plots data
set\footnote{https://www.kaggle.com/jrobischon/wikipedia-movie-plots}. What we
do here is to select all the classes whose support is above a certain threshold.
We then repeat the training with increasing values of the threshold to gradually
reduce the number of classes that the models see. The training sets are then
unbalanced at every run. The result of the experiment is summarised in Figure
\ref{wiki}. Again, we move from left to right as the threshold increases (and
the number of classes decreases). Contrary to the other two, the complement
model shows an upward trend, but as the accuracy grows the confidence attains
pretty low values.

\begin{figure}[t!]
  \begin{center}\small
  \scalebox{1}{\input{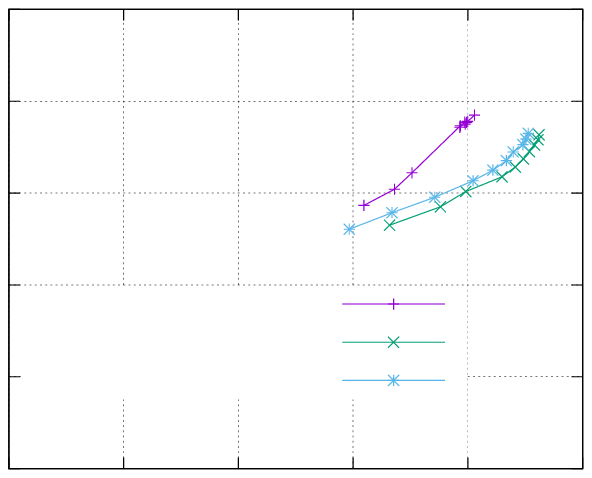}}
  \end{center}
  \caption{\label{ng20_balanced_purity}The accuracy \emph{versus} purity for the
  20 Newsgroup experiment with balanced supports.}
\end{figure}

\begin{figure}[t!]
  \begin{center}\small
  \scalebox{1}{\input{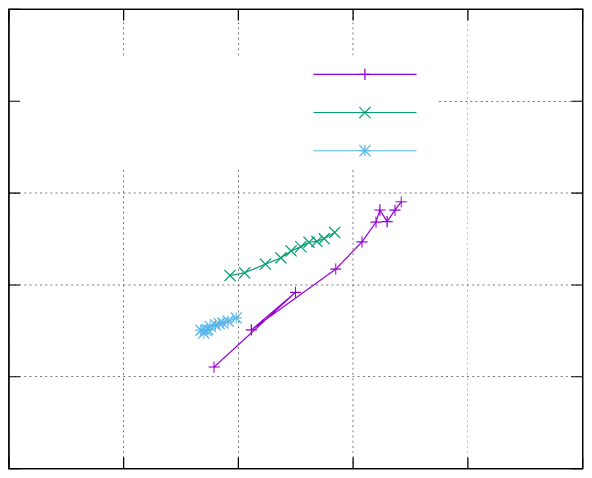}}
  \end{center}
  \caption{\label{ng20_unbalanced_purity}The accuracy \emph{versus} purity for
  the 20 Newsgroup experiment with unbalanced supports.}
\end{figure}

\begin{figure}[t!]
  \begin{center}\small
  \scalebox{1}{\input{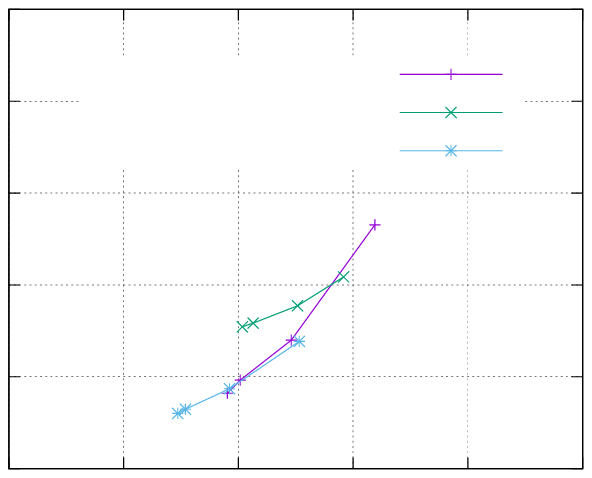}}
  \end{center}
  \caption{\label{wiki_purity}The accuracy \emph{versus} purity for the
  Wikipedia Movie Plots experiment.}
\end{figure}

For completeness, we present the results of accuracy \emph{vs} purity. Based on
the analysis of the behaviour of purity as an estimator of confidence, we expect
that, as the accuracy grows, the values of purity move from bottom to top.
Models that are very confident of the wrong answers will tend to live in the
bottom left corner of the plot, whereas models that make good and confident
decisions will be located in the top right corner. In the middle we have a
horizontal strip, between $0.5$ and $0.7$ of poorly confident models. We can see
from the three plots \ref{ng20_balanced_purity}, \ref{ng20_unbalanced_purity}
and \ref{wiki_purity} that the complement models are, as we expected, all
located near this middle strip.

\section{Concluding remarks}

It emerges from existing literature that the idea of using entropy to estimate
the confidence (or uncertainty) of a probabilistic classifier is, without any
doubts, natural and, we could add, almost common sense. Whilst we present the
metric \eqref{eq:entropy_score} as \emph{new}, we have also observed that, in
fact, many authors mention entropy or information when dealing with uncertainty
(see, e.g., the already cited \cite{zhang2019}). What we have done here was to
formalise this idea into the form of a $[0-1]$-metric that can give an estimate
of the average confidence of the predictions of a probabilistic classification
model.

The applications that we presented showed the usefulness of these metrics both
from a practical and a theoretical point of view. Indeed, we made use of the the
entropy considerations that are at the heart of the entropy score
\eqref{eq:entropy_score} to provide a theoretical explanation of the degraded
confidence observed with the Complement Na\"ive Bayes model of \cite{rennie}. On
the practical side, we have shown how the accuracy \emph{vs} entropy score plots
can provide a better insight into probabilistic classifiers, especially when
decision rules are based on a probability or score threshold. We believe that
being able to identify models that are good both in accuracy and confidence is
important in those applications where decisions need to be taken only when their
confidence is relatively high. For example, one might find that a Multinomial
Na\"ive Bayes model on an unbalanced data set provides satisfactory results in
terms of accuracy, and that a decision is taken based on its prediction provided
that the probability of the predicted class is above, say, 70\%. It is natural
to try different models to improve on the accuracy as much as possible, and the
Complement Na\"ive Bayes model of \cite{rennie} is one way to go. However, one
might then find that all the predictions do not go above, say, 30\% confidence,
meaning that, whilst the model yields higher accuracy, no decision will ever be
taken based in its predictions, rendering the ``better'' model
useless\footnote{One main point of objection is that, in many applications, deep
models tend to outperform shallow models. However, there might be hardware
constraints that make the training and/or deployment of deep models impossible,
and therefore one is limited to dealing with shallow models only.}.

In this paper we have focused on Na\"ive Bayes
models for the sole purpose of providing a good baseline for comparison against
the Complement Na\"ive Bayes model. Of course, this analysis can be carried out
with any probabilistic classifier.

\section{Acknowledgements}

The author wishes to thank Craig Alexander, Vinny Davies and Martina Pugliese
for their valuable comments and suggestions on an earlier draft of this paper.

\bibliography{bib}{}

\begin{thebibliography}{10}

\bibitem{domingos99}
Pedro Domingos.
\newblock {MetaCost}: A general method for making classifiers cost-sensitive.
\newblock In {\em Proceedings of the Fifth ACM SIGKDD International Conference
  on Knowledge Discovery and Data Mining}, KDD '99, pages 155--164, New York,
  NY, USA, 1999. Association for Computing Machinery.

\bibitem{fawcett06}
Tom Fawcett.
\newblock An introduction to {ROC} analysis.
\newblock {\em Pattern Recogn. Lett.}, 27(8):861--874, 2006.

\bibitem{scikit-learn}
F.~Pedregosa, G.~Varoquaux, A.~Gramfort, V.~Michel, B.~Thirion, O.~Grisel,
  M.~Blondel, P.~Prettenhofer, R.~Weiss, V.~Dubourg, J.~Vanderplas, A.~Passos,
  D.~Cournapeau, M.~Brucher, M.~Perrot, and E.~Duchesnay.
\newblock Scikit-learn: Machine learning in {P}ython.
\newblock {\em Journal of Machine Learning Research}, 12:2825--2830, 2011.

\bibitem{rennie}
Jason D.~M. Rennie, Lawrence Shih, Jaime Teevan, and David~R. Karger.
\newblock Tackling the poor assumptions of naive bayes text classifiers.
\newblock In {\em Proceedings of the Twentieth International Conference on
  International Conference on Machine Learning}, ICML'03, pages 616--623. AAAI
  Press, 2003.

\bibitem{ritschard06}
Gilbert Ritschard.
\newblock Computing and using the deviance with classification trees.
\newblock In Alfredo Rizzi and Maurizio Vichi, editors, {\em Compstat 2006 -
  Proceedings in Computational Statistics}, pages 55--66, Heidelberg, 2006.
  Physica-Verlag HD.

\bibitem{schetinin04}
Vitaly Schetinin, Derek Partridge, Wojtek~J. Krzanowski, Richard~M. Everson,
  Jonathan~E. Fieldsend, Trevor~C. Bailey, and Adolfo Hernandez.
\newblock Experimental comparison of classification uncertainty for randomised
  and bayesian decision tree ensembles.
\newblock In Zheng~Rong Yang, Hujun Yin, and Richard~M. Everson, editors, {\em
  Intelligent Data Engineering and Automated Learning -- IDEAL 2004}, pages
  726--732, Berlin, Heidelberg, 2004. Springer Berlin Heidelberg.

\bibitem{spackman89}
Kent~A. Spackman.
\newblock Signal detection theory: Valuable tools for evaluating inductive
  learning.
\newblock In {\em Proceedings of the Sixth International Workshop on Machine
  Learning}, pages 160--163, San Francisco, CA, USA, 1989. Morgan Kaufmann
  Publishers Inc.

\bibitem{stehman97}
Stephen~V. Stehman.
\newblock Selecting and interpreting measures of thematic classification
  accuracy.
\newblock {\em Remote Sensing of Environment}, 62(1):77--89, 1997.

\bibitem{tibshirani96}
Robert Tibshirani.
\newblock {\em Bias, variance and prediction error for classification rules}.
\newblock University of Toronto, Department of Statistics, 1996.

\bibitem{cen}
Xiao-Ning Wang, Jin-Mao Wei, Han Jin, Gang Yu, and Hai-Wei Zhang.
\newblock Probabilistic confusion entropy for evaluating classifiers.
\newblock {\em Entropy}, 15(12):4969--4992, Nov 2013.

\bibitem{zhang2019}
Xuchao Zhang, Fanglan Chen, Chang-Tien Lu, and Naren Ramakrishnan.
\newblock Mitigating uncertainty in document classification.
\newblock In {\em Proceedings of the 2019 Conference of the North {A}merican
  Chapter of the Association for Computational Linguistics: Human Language
  Technologies, Volume 1 (Long and Short Papers)}, pages 3126--3136,
  Minneapolis, Minnesota, 2019. Association for Computational Linguistics.

\end{thebibliography}
\bibliographystyle{plain}

\end{document}